# Cyberbotics Ltd.
# Webots$^{TM}$: Professional Mobile Robot Simulation

**Olivier Michel**
Cyberbotics Ltd.,
Swiss Federal Institute of Technology in Lausanne, BIRG & SWIS research groups,
Olivier.Michel@cyberbotics.com

***Abstract:*** *Cyberbotics Ltd. develops Webots$^{TM}$, a mobile robotics simulation software that provides you with a rapid prototyping environment for modelling, programming and simulating mobile robots. The provided robot libraries enable you to transfer your control programs to several commercially available real mobile robots. Webots$^{TM}$ lets you define and modify a complete mobile robotics setup, even several different robots sharing the same environment. For each object, you can define a number of properties, such as shape, color, texture, mass, friction, etc. You can equip each robot with a large number of available sensors and actuators. You can program these robots using your favorite development environment, simulate them and optionally transfer the resulting programs onto your real robots. Webots$^{TM}$ has been developed in collaboration with the Swiss Federal Institute of Technology in Lausanne, thoroughly tested, well documented and continuously maintained for over 7 years. It is now the main commercial product available from Cyberbotics Ltd.*

## 1. Introduction

Cyberbotics Ltd. was founded in 1998 as a spin-off company from the Swiss Federal Institute of Technology in Lausanne (EPFL). It currently employs two people and develops Webots$^{TM}$: a commercial software used for mobile robotics prototyping simulation and transfer to real robots (see Fig. 1). In 1998 and 1999 Cyberbotics developed an Aibo$^®$ simulator for Sony Ltd. Cyberbotics has now collaborations with the Biologically Inspired Robotics Group (BIRG) and the Swarm Intelligent System Research Group (SWIS) of the EPFL through the Swiss CTI technology transfer program.
Webots$^{TM}$ runs on Windows, Linux and Mac OS X and is intended for researchers and teachers interested in mobile robotics. It is commercially available from Cyberbotics Ltd. (http://www.cyberbotics.com).

## 2. Need for Simulation

Although the final aim is real robotics, it is often very useful to perform simulations prior to investigations with real robots. This is because simulations are easier to setup, less expensive, faster and more convenient to use. Building up new robot models and setting up experiments only takes a few hours. A simulated robotics setup is less expensive than real robots and real world setups,thus allowing a better design exploration. Simulation often runs faster than real robots while all the parameters are easildisplayed on screen. Simulations make it possible to use computer expensive algorithms that would need ages to run on real robot micro-controllers, like genetic algorithms. Finally, the simulation results are transferable onto the real robots.

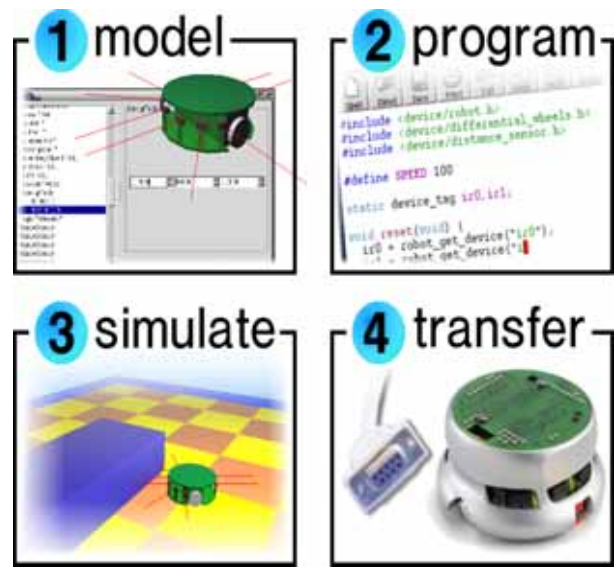

Fig.1. Stages of development of a robot simulation

## 3. Features

Webots[TM] has a number of essential features intended to make this simulation tool both easy to use and powerful:
- Models and simulates any mobile robot, including wheeled, legged and flying robots.
- Includes a complete library of sensors and actuators.
- Lets you program the robots in C, C++ and Java, or from third party software through TCP/IP.
- Transfers controllers to real mobile robots, including Aibo®, Lego® Mindstorms®, Khepera®, Koala® and Hemisson®.
- Uses the ODE (Open Dynamics Engine) library for accurate physics simulation.
- Creates AVI or MPEG simulation movies for web and public presentations.
- Includes many examples with controller source code and models of commercially available robots.
- Lets you simulate multi-agent systems, with global and local communication facilities.

## 4. Robot and world editor

A library of sensors is provided so that you can plug a sensor in your robot model and tune it individually (range, noise, response, field of view, etc.). This sensor library includes distance sensors (infra-red and ultra-sonic), range finders, light sensors, touch sensors, global positioning sensor (GPS), inclinometers, compass, cameras (1D, 2D, color, black and white), receivers (radio and infra-red), position sensors for servos, incremental encoders for wheels.

Similarly, an actuator library is provided. It includes differential wheel motor unit, independent wheel motors,

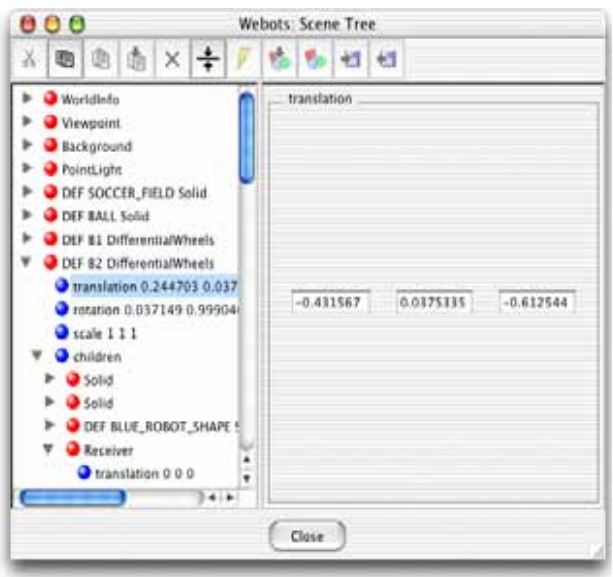

Fig. 2. Scene tree editor window

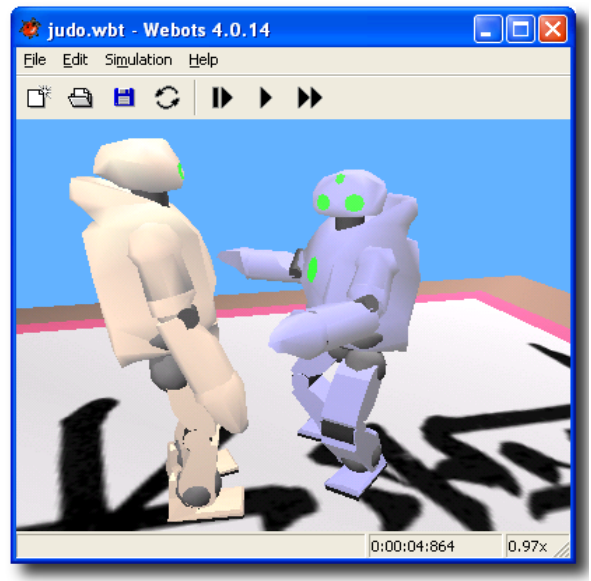

Fig.3. Humanoid robot modelling using physics

servos (for legs, arms, etc.), LEDs, emitters (radio and infra-red) and grippers.

With Webots[TM], you can create complex environments for your mobile robot simulations, using advanced hardware accelerated OpenGL technologies, including lighting, smooth shading, texture mapping, fog, etc. Moreover, Webots[TM] allows you to import 3D models in its scene tree (see Fig. 2) from most 3D modelling software through the VRML97 standard.

You can create worlds as large as you need and Webots[TM] will optimize them to enable fast simulations. Complex robots can be built by assembling chains of servo nodes. This allows you to easily create legged robots with several joints per leg (as shown in Fig. 3), robot arms, pan / tilt camera systems, etc. For example, you can place several cameras on the same robot to perform binocular stereo vision, or 360 degree vision systems.

## 5. Realistic Simulation

The simulation system used in Webots[TM] uses virtual time, making it possible to run simulations much faster than it would take on a real robot. Depending on the complexity of the setup and the power of your computer, simulations can run up to 300 times faster than the real robot when using the fast simulation mode. The basic simulation time step can be adjusted to suit your needs (precision versus speed). A step-by-step mode is available to study in detail how your robots behave

Simulating complex robotic devices including articulated mechanical parts requires precise physics simulation. Webots[TM] relies on ODE (Open Dynamics Engine) to perform accurate physics simulation wherever it is necessar y

For each component of a robot, you can specify a mass distribution matrix (or use primitives for simple geometries), static and kinematic friction coefficients, bounciness, etc. Moreover each component is associated with a bounding object used for collision detection. Servo devices can be controlled by your program in torque, position or velocity. The control parameters for the servo can be individually adjusted from your controller program.

The graphical user interface of Webots$^{TM}$ allows you to easily interact with the simulation while it is running. By dragging the mouse, you can change the viewpoint position, orientation and zoom using the mouse wheel. Pressing the shift key while dragging the mouse allows you to move or rotate objects. This feature facilitates interactive testing.

### 6. Programming Interface

Programming your robot using the C language is as simple as shown in Fig. 4.

In this example, the robot is a differential wheeled robot equipped with an infra-red distance sensor named "ir" looking forward. The robot will stop moving if the distance sensor detects an obstacle and restart moving when the obstacle is no longer detected. A similar Java programming interface is also included. Moreover, any Webots$^{TM}$ controller can be connected to a third party software program, such as MatLab®, LabView®, Lisp®, etc. through a TCP/IP interface. Research experiments often need to interact automatically with the simulation. The supervisor capability allows you to write a program responsible for supervising the experiment. Such a program can dynamically move objects, send messages to robots, record robot trajectories, add new objects or robots, etc.

The supervisor capability can be used in computationally expensive simulations where a large number of robot configurations and control parameters have to be evaluated, as in genetic evolution, neural networking, machine learning, etc.

```
#include <robot.h>
#include <differential_wheels.h>
#include <distance_sensor.h>

DeviceTag ir;

void my_robot_reset() {
  ir = robot_get_device("ir");
}

void main() {
  robot_live(my_robot_reset);
  for(;;) { /* infinite loop */
    if (distance_sensor_get_value(ir)>100)
      differential_wheels_set_speed(0,0);
    else
      differential_wheels_set_speed(10,10);
    robot_step(64) /* run for 64 ms */
  }
}
```

Fig. 4. Programming a robot with the C interface

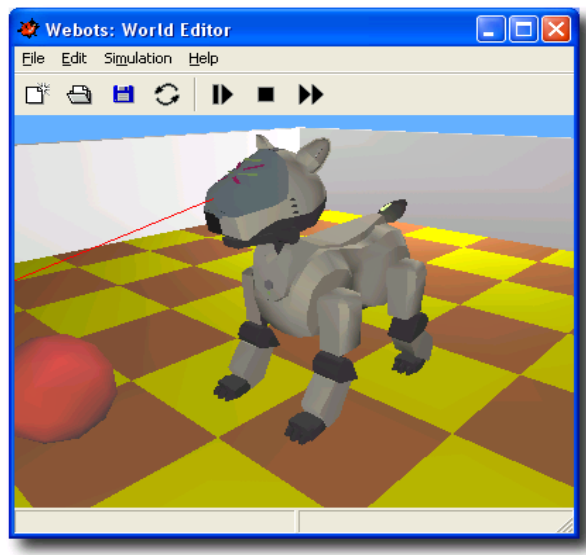

Fig. 5. Aibo ERS-210 robot with transfer capability

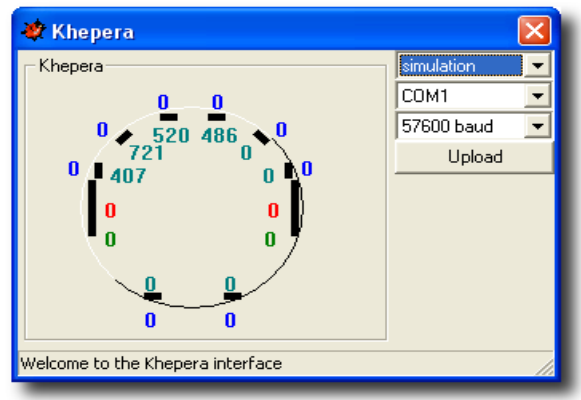

Fig. 6. Control for a real and simulated Khepera robot

### 7. Transfer to real robots

Once tested in simulation your robot controllers can be transferred to real robots:
- Khepera® and Koala®: cross-compilation of C Webots$^{TM}$ controllers and remote control with any programming language (see Fig. 6).
- Hemisson®: Finite state automata graphical programming with remote control and autonomous execution modes.
- LEGO® Mindstorms®: cross-compilation for RCX of Java Webots$^{TM}$ controllers based on LeJOS.
- Aibo®: cross-compilation of C/C++ Webots$^{TM}$ controller programs based on the Open-R SDK (see Fig. 5).
- Your own robot: The Webots$^{TM}$ user guide explains how to build your own Webots$^{TM}$ cross-compilation system for your very own robot.

### 8. Documentation and support

Webots$^{TM}$ comes with a complete documentation, including two printed manuals with color covers. This documentation is also included on the Webots$^{TM}$ CD-ROM in both PDF and HTML format.

The Webots[TM] user guide explains how to install and get started with Webots[TM]. This manual includes a step-by-step tutorial for modelling and programming your own robot, and describes a number of sample experiments included on the CD-ROM. It explains the basic principles of Webots[TM] and shows you how to transfer your programs to real robots. The Webots[TM] reference manual contains everything you need to develop your Webots[TM] application. It provides a complete description of all the objects you can simulate with Webots[TM], including robot bases, sensors, actuators, simple objects, etc. The programming interface is completely documented with examples. Functions are sorted by categories. Finally a number of example worlds and controllers are provided on the CD-ROM which can serve as a starting point for developing your application. Webots[TM] users take advantage of the Webots[TM] users community through a support mailing list. Most questions are answered within 24 hours by Cyberbotics support services. All Webots[TM] licenses include one year of personalized user support and free upgrades via the Internet.

## 9. Examples of Applications

Webots[TM] has been used by more than one hundred universities and research centers worldwide since 1998

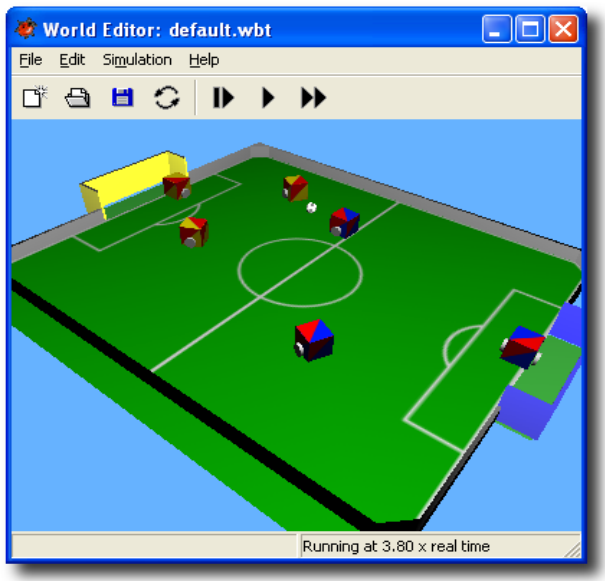

Fig. 7. Multi-agent robot soccer simulation

for both education and research purposes.
Education applications allow the students to get started with robotics, 3D modelling, programming, artificial intelligence, computer vision, artificial life, etc. with an integrated tool. It is easy to implement a virtual robot contest, like a soccer contest or a humanoid locomotion contest, based on Webots[TM], which is highly motivating for the students.
All the articles listed in the last section of this paper refer to different research applications of Webots[TM]. It would be out the scope of this paper to present each application. However, we may attempt to classify them into several major categories:

- The multi-agent simulations category includes research experiments where several robots cooperate to reach a global goal (see Fig. 7).
- The artificial intelligence category attempts to validate psychology hypothesis, mainly learning, by simulating intelligent mobile robot behaviors.
- The control research category involves developing efficient control algorithms to perform complex mobile robot motion.
- The robot design category aims at shaping mobile robots defining the position and properties of its sensors and actuators. This is especially useful to investigate new wheeled, legged or flying robots.